\documentclass[conference]{IEEEtran}
\usepackage[utf8]{inputenc}
\usepackage[T1]{fontenc}

\IEEEoverridecommandlockouts
\usepackage{cite}
\usepackage{amsmath,amssymb,amsfonts}
\usepackage{algorithmic}
\usepackage{graphicx}
\usepackage{textcomp}
\usepackage[most]{tcolorbox}
\usepackage{xcolor}
\usepackage{amsthm}
\newtheorem{definition}{Definition}
\usepackage[T1]{fontenc}
\usepackage{xspace}
\usepackage[utf8]{inputenc}
\usepackage[T1]{fontenc}
\usepackage{graphicx}
\usepackage{longtable}
\usepackage{booktabs}
\usepackage{tabularx}
\usepackage[hidelinks]{hyperref}
\usepackage{array}
\usepackage{multirow}
\usepackage{amssymb}
\usepackage[figuresright]{rotating}
\usepackage{amsmath}
\usepackage{subcaption}
\usepackage{framed}
\usepackage{hyperref}
\usepackage{enumitem}

\newcommand{\q}[1]{``#1''}
\newcommand{\DR}{\textsf{Disturbing Rule}\xspace}
\newcommand{\DRs}{\textsf{Disturbing Rule(s)}\xspace}

\newcommand{\DI}{\textsf{Device of Interest}\xspace}
\newcommand{\CS }{\textsf{Confusing Situation}\xspace}
\newcommand{\CST}{\textsf{Current State}\xspace}
\newcommand{\PST}{\textsf{Previous State}\xspace}
\newcommand{\EST}{\textsf{Expected State}\xspace}

\newcommand{\AR}{\textsf{Appropriate Rule}\xspace}
\newcommand{\ARs}{\textsf{Appropriate Rule(s)}\xspace}

\newcommand{\exci}{\textsf{Case E1}\xspace}
\newcommand{\excii}{\textsf{Case E2}\xspace}
\newcommand{\exciii}{\textsf{Case E3}\xspace}

\newcommand{\foili}{\textsf{Case  F1}\xspace}

\newcommand{\foil}{\emph{Foil}\xspace}

\def\BibTeX{{\rm B\kern-.05em{\sc i\kern-.025em b}\kern-.08em
    T\kern-.1667em\lower.7ex\hbox{E}\kern-.125emX}}
\begin{document}

% Avoid orphans and widows
\clubpenalty = 10000
\widowpenalty = 10000
\displaywidowpenalty = 10000

\title{From Facts to Foils: Designing and Evaluating Counterfactual Explanations for Smart Environments}

\author{
\IEEEauthorblockN{Anna Trapp}
\IEEEauthorblockA{University of Cologne\\ Cologne, Germany \\ atrapp2@smail.uni-koeln.de}
\and
\IEEEauthorblockN{Mersedeh Sadeghi*\thanks{*Corresponding author: mersedeh.sadeghi@uni-koeln.de}}
\IEEEauthorblockA{University of Cologne\\ Cologne, Germany\\
mersedeh.sadeghi@uni-koeln.de}
\and
\IEEEauthorblockN{Andreas Vogelsang}
\IEEEauthorblockA{paluno - The Ruhr Institute for Software Technology \\
University of Duisburg-Essen, Essen, Germany \\
andreas.vogelsang@uni-due.de}
}

\maketitle

\begin{abstract}
Explainability is increasingly seen as an essential feature of rule-based smart environments. While counterfactual explanations, which describe what could have been done differently to achieve a desired outcome, are a powerful tool in eXplainable AI (XAI), no established methods exist for generating them in these rule-based domains. In this paper, we present the first formalization and implementation of counterfactual explanations tailored to this domain. It is implemented as a plugin that extends an existing  explanation engine for smart environments. We conducted a user study (N=17) to evaluate our generated counterfactuals against traditional causal explanations. The results show that user preference is highly contextual: causal explanations are favored for their linguistic simplicity and in time-pressured situations, while counterfactuals are preferred for their actionable content, particularly when a user wants to resolve a problem. Our work contributes a practical framework for a new type of explanation in smart environments and provides empirical evidence to guide the choice of when each explanation type is most effective.
\end{abstract}


\section{Introduction}
\label{sec: Introduction}
Smart environments, such as smart homes, offices, and buildings, integrate sensor-enabled devices to support users in decision-making, monitoring, and managing abnormal situations~\cite{eldin2021smartenvironments, ahmed2016stateoftheart}. The rapid adoption of these environments is fueled by advances in the Internet of Things (IoT) and Artificial Intelligence (AI), decreasing device costs, and improved system integration~\cite{li2021motivations, baresi2018tdex, baresi2018fine}. 

Rule-based systems are a prevalent approach for implementing automation in smart environments, by executing predefined rules when certain conditions are met~\cite{nandi2016RBS, sadeghi2024smartex}. Despite their prevalence, this automation can be difficult for users to interpret. This problem is exacerbated in multi-user environments where rules are created and managed jointly, as users frequently struggle to understand the internal logic driving a particular action or system response.

Providing explanations has been shown to significantly enhance understanding, user perception, and task performance by offering insights into system behavior and highlighting causal factors~\cite{chazette2020requirement}. Explainability is also strongly linked to trust and transparency: mismatches between user expectations and system behavior erode trust, whereas explanations can mitigate this effect~\cite{lim2009expintelligibility, sadeghi2021cases, winikoff2018trust, sadeghi2024explaining}. Consequently, explainability is now increasingly integrated into a wide range of intelligent systems and application domains~\cite{sakai2022explainable,unterbusch2023explanation,yeong2025exploring}.

Among various explanation types, counterfactual explanations are particularly promising. They explain an outcome by describing what would need to have been different for an alternative outcome to have occurred (e.g., \q{A would have happened if...})~\cite{stepin2021survey}. This form of reasoning is deeply intuitive, mirroring how humans learn and infer causality~\cite{guidotti2022counterfactualreview, byrne2019CFEinXAI}. Counterfactual explanations are uniquely suited for smart environments for several reasons. First, they excel in controllable and repeatable situations where users want to learn corrective actions~\cite{roese1997counterfactualthinking}. Second, they directly answer the \q{how to} and \q{what if} questions that are most useful in proactive systems, rather than just post-hoc \q{why} questions~\cite{lim2009expintelligibility, woodward2005theory}. Finally, by showing users how to achieve a desired result, they align with the principles of the GDPR's \q{Right to Explanation} without needing to expose the entire internal logic of the system~\cite{wachter2017blackbox}.

Despite this strong potential, there is currently no formal definition of counterfactual explanations for rule-based smart environments, nor are there established methods for their generation. This paper addresses this research gap with the following contributions:
\begin{enumerate}
\item We propose a formal, literature-grounded definition of counterfactual explanations tailored to the context of rule-based smart environments.
\item We present a novel framework for generating these explanations and provide an implementation to demonstrate its feasibility.
\item We conduct a user-centric evaluation of our approach, addressing a significant need for more user studies in the field of counterfactual research~\cite{guidotti2022counterfactualreview}.
\end{enumerate}

%%%%%%%%%%%%%%%  Related Works  %%%%%%%%%%%%%%%

\section{Background and Related Work}
\label{sec: Background and Related Work} 
Automation is reshaping software engineering, replacing manual tasks with intelligent processes across the development lifecycle and application domains~\cite{ryalat2023design,kalwar2021smart, hosseini2019automated}. Smart homes exemplify this shift, highlighting the increasing role of automation in everyday life.
Smart environments are sensor-driven systems that autonomously perceive, reason, and act to enhance user comfort~\cite{ahmed2016stateoftheart, alam2012smarthomes}. A common implementation strategy involves \textit{rule-based systems}~\cite{nandi2016RBS}, which consist of a knowledge base (rules and device states) and an inference engine that evaluates rule preconditions and fires rules when satisfied~\cite{masri2019surveyrbs}.

Each rule comprises logical \textit{preconditions} and \textit{actions}. When multiple rules with conflicting actions are simultaneously eligible, a conflict resolution strategy is required. While approaches like specificity or recency exist~\cite{ali2018conflictresolution}, we adopt \textit{priority-based scheduling}~\cite{shah2019conflict}, where each rule is assigned a unique priority, and the highest-priority rule is executed.

To support transparency, smart environments can incorporate an \textit{explanation layer}~\cite{masri2019surveyrbs}. A system is considered explainable if it provides information that enables users to understand specific outcomes~\cite{madumal2020rl}. While causal explanations describe how system logic led to a decision, \textit{counterfactual explanations} describe how an alternative condition could have produced a different outcome~\cite{wachter2017blackbox}.

Theoretical foundations for counterfactual reasoning include the \textit{possible worlds} view~\cite{lewis1973counterfactuals}, \textit{structural causal models}~\cite{pearl2000scm}, and \textit{interventionist accounts}~\cite{woodward2005theory}, all of which inform definitions used in XAI. Most definitions emphasize \textit{minimality}—changing as little as possible to achieve a different result~\cite{guidotti2022counterfactualreview}. In our setting, we define counterfactual explanations as the minimal change to explanation constructs required to achieve a desired alternative outcome.

\paragraph{Explainability in Smart Environments.}
Several frameworks enable explainability in IoT contexts. \textit{MAB-EX} supports runtime explanation generation for cyber-physical systems~\cite{blumreiter2019selfexplainable}, while a modular architecture by~\cite{houze2022architecture} introduces Local Explanatory Components (LECs) for per-device transparency. Agent-based systems have also been deployed in lab environments~\cite{dobrovolskis2023agentbased}. Explainable human activity recognition has been applied in caregiver monitoring systems~\cite{das2023activityrecognition}.

\textit{SmartEx}~\cite{sadeghi2024smartex} provides context-aware explanations in rule-based environments and was extended by~\cite{herbold2024contrastive} to support contrastive reasoning.

\paragraph{Counterfactual Explanations in XAI.}
In XAI, counterfactuals are used to show how minimal input changes lead to different predictions. Several methods ensure feasibility or diversity: \textit{FACE} emphasizes realistic paths, \textit{DiCE} focuses on generating diverse outputs, and \textit{FOCUS} handles tree-based models~\cite{lucic2022focus,mothilal2020MLClassifiers}.Other work uses GANs, ASP, or interpretable decision trees~\cite{Waa2018localfoiltrees,bertossi2020asp}. Counterfactuals have also been applied to recommender systems and reinforcement-learned causal models~\cite{madumal2020rl}.

Despite this broad literature, to our knowledge, no prior work has addressed counterfactual explanations for rule-based smart environments. Our work closes this gap by introducing a formal definition and a practical generation framework tailored to this class of systems.

%\section{Background}\label{sec:back}

%To contextualize these definitions, consider the following example. Alice is lying on the couch in the living room, reading a book, when the overhead stand light (the \DI) suddenly turns off. This observed event constitutes the fact and defines the \CST of the \DI as \q{off}. This surprises her, as she expects the light to remain on (the foil) based on the Rule-A she configured to keep the light on as long as someone are in the living room. This expectation defines the \EST as \q{on}. The state of the light immediately before this event was also \q{on}, making this the \PST.

%It is important to note that in the context of our work, concepts like \textit{Rule Activation/Inactivation} are presented as explanations to the user. They describe the concrete actions that a user, or the system itself, could perform to achieve a desired outcome by manipulating which rules are active or fired.
\section{Approach}
\label{sec: Approach} 

\begin{table*}[t]
\centering
\caption{Core definitions used in this work.}
\label{tab:defs}
\renewcommand{\arraystretch}{1.12}
\begin{tabularx}{\textwidth}{@{}l X@{}}
\toprule
\textbf{Term} & \textbf{Definition} \\
\midrule
\textbf{Confusing Situation} &
A \CS arises when there is a discrepancy between the observed reality (the \emph{Fact}) and the user’s expectation (the \emph{Foil}). \\ \midrule

\textbf{Device of Interest} &
The \DI is the specific device responsible for the \CS, i.e., the component of the smart environment whose state contradicts user expectations. \\ \midrule

\textbf{\DI States:} & \\
\textbf{Previous State} & The state of the \DI at an initial time $t_0$.\\
\textbf{Current State} & The state of the \DI at a subsequent time $t_1$, when the need for an explanation arises.\\
\textbf{Expected State} & The state the user anticipated for the \DI, derived from the foil; used interchangeably with \textit{Foil} in this paper. 

%=====>>>>>> While \CST and \PST are observable from the execution history, \EST is inferred by the explanation engine \cite{herbold2024contrastive}. A basic assumption for any \CS is \CST $\neq$ \EST; \PST may coincide with either. 
\\ \midrule

\textbf{Disturbing Rules} &
The set of \DR comprises all rules whose preconditions are satisfied (true) in the current system state. The presence of these enabled rules prevents the system from achieving or maintaining the \emph{Foil}. A conflict–resolution mechanism (e.g., priority) selects one such rule to execute; its execution directly causes the state transition leading to the \CS. \\ \midrule

\textbf{Appropriate Rule} &
An \AR is a rule whose actions can bring the \DI to the desired state. An \AR may itself be \textit{active} or \textit{inactive}. \\ \midrule

\textbf{Rule State (Active / Inactive)} &
A rule is \textit{active} if all of its preconditions hold in the current system state; otherwise, it is \textit{inactive}. \\ \midrule

\textbf{Rule Priority} &
A property of each rule used for conflict resolution, ensuring that when multiple rules are active simultaneously, only the one with the highest priority is fired. \\ \midrule

\textbf{Rule Activation} &
The actions required to satisfy all preconditions of a rule, thereby making it active. \\ \midrule

\textbf{Rule Inactivation} &
Preventing a rule from being fired, which is achieved by making at least one true precondition false.
\\\midrule

\textbf{Overriding a Rule} & Preventing a rule from being fired, by activating another rule with a higher priority. 
\\

\bottomrule
\end{tabularx}
\end{table*}
We define counterfactual explanations for rule-based smart environments by extending established notions of minimal change~\cite{guidotti2022counterfactualreview, bertossi2020asp}. Table~\ref{tab:defs} lists the core concepts and terminology that underpin our proposed framework.

\begin{definition}A counterfactual explanation in a smart environment is an explanation containing the minimal change to explanation constructs, such that a specific \foil would have occurred instead of the \emph{Fact}. \label{def CFE}
\end{definition}

\textit{Explanation Constructs} refer to specifications, facts, propositions, and events relating to both system internals and the external environment~\cite{sadeghi2024smartex}. Following contrastive explanation theory~\cite{lipton1990contrastive, miller2021contrastive}, we define the \textbf{Fact} as the actual outcome prompting an explanation request and the \textbf{Foil} as the user-expected alternative outcome. 

In the context of smart environments, a user requiring an explanation is typically not interested in any possible alternative to an event, but rather in achieving a specific, desired outcome. This user-centric requirement informs our choice of a counterfactual framework. While many theories ~\cite{guidotti2022counterfactualreview, bertossi2020asp} define the \foil broadly as any event differing from the \emph{Fact}, we align our methodology with that of Wachter et al. ~\cite{wachter2017blackbox}, which defines the \foil as a single, specific incident. This approach ensures the generated explanation is directed toward achieving the user's explicit goal. Within this scope, our framework is concerned solely with constructing an explanation why the observed outcome occurred instead of the specific \foil provided by the framework proposed by Herbold et al. ~\cite{herbold2024contrastive}; we do not evaluate the validity of the \foil itself. The process, therefore, relies on the assumption that this \foil accurately represents the user’s intended outcome.

Lastly, Counterfactual explanations can be \textbf{Additive}  or \textbf{Subtractive}  ~\cite{roese2017theory}. In rule-based system, additive explanations aim to fire \ARs (Table~\ref{tab:defs}) to reach the \foil, a process associated with creative problem-solving. In contrast, subtractive explanations work by preventing a \DR (Table~\ref{tab:defs}) from activating, which supports analytical reasoning.~\cite{markman2007addsub}.

\subsection { Cases for Explanation Needs}\label{sub:ExCases}
We identify three distinct cases of explanation needs for counterfactual explanations in smart environments.

\textbf{Case E1: An Undesired Event Occurred.}
The first type of \CS(Table~\ref{tab:defs}) occurs when, at an initial time $t_0$, the system behaves as expected. At a subsequent time $t_1$, however, a \DR is triggered, causing the \DI(Table~\ref{tab:defs}) to transition to an unexpected state. In this case, the \emph{Foil} corresponds to maintaining the state that existed at $t_0$. Hence, $\EST = \PST$, while the \CST is distinct (Table~\ref{tab:defs}).

\textbf{Case E2: An Expected Event Did Not Occur.}
This case describes confusion due to system inaction, where the state remains unchanged from $t_0$
to $t_1$ ($\PST = \CST$). The \emph{Fact} is this persistence, while the \emph{Foil} is the transition to the very state of the \DI, that the user anticipated, i.e., \EST.

\textbf{Case E3: A Different Event Occurred.}
This case represents a \CS where an event occurred (A \DR fired) and brought the \DI into a particular state. So, the system has transitioned from a \PST at time $t_0$ to a new $\CST$ (the \emph{Fact}) at time $t_1$, but the user expected a transition to a different state altogether (the \emph{Foil}). Consequently, all three states, \PST, \CST, and \EST, are distinct from one another.

\subsection{Foil Achievement Strategy Selection.}
Counterfactual explanations focus on \q{contrary-to-fact} reasoning: they describe how a desired outcome (\foil) could have been achieved by analyzing what would need to differ from the current situation ~\cite{stepin2021survey}. In our framework, this involves first understanding why the \EST did not occur, identifying what must change to achieve it, and then describing such changes in the form of an explanation.

Table~\ref{tab:foil_achievement} summarizes the three general strategies for foil achievement in a rule-based system, which depend on the presence or absence of \ARs and \DRs. In short, the \foil may be realized by (i) \textit{Activating} (and ultimately firing, see Table~\ref{tab:defs}) some \ARs, 
 (ii) \textit{Inactivating} (Table~\ref{tab:defs}) \DRs that preempt \AR, or (iii)
a combination of both.

Each Explanation Case (\exci--\exciii) can map to one or more of these strategies, depending on whether the system lacked active rules, contained preempted rules, or was constrained by reinforcing \DR.\footnote{For instance, the \excii has three sub-cases, each mapping to a different one of the three strategies based on the underlying cause of the system's inaction. However, due to space constraints, we omit a detailed analysis of these mappings in this manuscript. The complete specification is available in our extended work at \url{http://kups.ub.uni-koeln.de/id/eprint/78813}.}
\begin{table*}
\centering
\caption{Foil-achievement cases: reasons for failure, required differences, and resolution strategies}
\label{tab:foil_achievement}
\begin{tabularx}{\textwidth}{c p{4.5cm} p{5.7cm} p{5.8cm}}
\toprule
\textbf{Case} & \textbf{Properties of Factual Situation} & \textbf{What Must Have Been Different to Achieve \emph{Foil}} & \textbf{What Must Be Done to Achieve the 
\emph{Foil}} \\
\midrule
F1 & No \AR was active, and no \DR was present. & At least one \AR should have been active. & Identify an \AR and \textit{Activate} it.\\\midrule

F2 & An \AR was active, but it was preempted by a \DR with a higher priority. & No \DR with a higher priority than the \AR should have been active. & \textit{Inactivate} all \DR that have a higher priority than the \AR.\\\midrule

F3 & No \AR was active, while at least one \DR was. & No \DR, but,  at least one \AR should have been active.&  All \DR must be inactivated and one \AR must be identified and activated \\
\bottomrule
\end{tabularx}
\end{table*}

\paragraph{Additive Counterfactual Explanation} it is a description of how to achieve the \EST through the \foili. In this case, all \ARs with a priority higher than or equal to the highest-priority \DR (if any) are considered. For each such \AR, the \emph{minimal set of changes} required to activate it is computed (See Section~\ref{Sec:minimal}). Since a rule fires only when \emph{all} of its preconditions are satisfied, each false precondition must be resolved. This can be achieved either directly, by modifying the state of relevant devices to satisfy the required preconditions, or indirectly, by firing other rule(s) whose actions enable those preconditions to be met. In the indirect case, the minimal changes needed to fire the supporting rule are determined recursively in the same manner. For each precondition, the framework selects the smaller of the direct or indirect cost. This process is repeated across all false preconditions of the candidate \AR, yielding the minimal change set required to activate it. Once this minimal change set is established for each candidate \AR, the framework selects the \AR with the overall smallest change set. The resulting \emph{Additive Counterfactual Explanation} consists of a description of these minimal changes, which represent the necessary actions to achieve the \foil.

\paragraph{Subtractive Counterfactual Explanations.}
It describes how to achieve the \foil by \textit{Inactivating} one or more \DRs. In this case, the minimal change required to falsify at least one precondition of the \DR is computed. For each precondition, the framework determines the modification that would make it false and selects the option requiring the smallest change. As in the additive case, such a modification can be implemented either directly or indirectly by firing another rule whose action changes the relevant state. The latter is again evaluated recursively in the same manner as for additive explanations.

An additional complication arises when a precondition to be falsified (e.g., \q{device $d$ is in state $s_1$}) is continuously reinforced by another rule $r_1$
with already satisfied preconditions, whose action enforces d→$s_1$. In such situations, falsifying the precondition requires not only altering the device state but also inactivating $r_1$ itself, since otherwise the system would immediately restore d to $s_1$. The resulting \emph{Subtractive Counterfactual Explanation} therefore provides a description of the minimal set of changes required to \textit{inactivate} the blocking \DR(s) and prevent their effects from recurring, thereby enabling the system to achieve the \foil.

\subsection{Minimal Change Computation}\label{Sec:minimal}

In line with research on human-centric explanations, our methodology follows the principle of \textit{minimal change}, as humans tend to prefer simple explanations that cite only main causes while ignoring unnecessary details~\cite{miller2019expAI, chazette2020requirement}. In the context of counterfactuals, this means providing the smallest set of actions and information required for the user to achieve their \foil. Accordingly, we identify five desirable properties that a counterfactual explanation should fulfil (see Table~\ref{tab:desirable_properties}), which our framework evaluates in a two-phase process.

\paragraph{Candidate Filtering with Controllability}
\textit{Controllability}, which measures the user's ability to enact a suggested change, is arguably the most critical property. From a practical standpoint, providing an explanation that relies on uncontrollable changes (e.g., weather conditions) is ineffective. Therefore, our framework uses \textit{Controllability} as a primary filter. Changes that are \textit{immutable} are generally discarded, while \textit{mutable but non-actionable} changes are evaluated by recursively analyzing the controllability of their underlying rule preconditions.

\paragraph{Scoring and Ranking with MCDM (Multi-Criteria Decision Making) Method}
Candidates that pass the Controllability filter are then scored and ranked based on four quantitative properties: \textit{Sparsity}, \textit{Temporality}, \textit{Proximity}, and \textit{Abnormality}. All four of these properties are conceptually non-beneficial measures, meaning that lower values indicate more desirable minimal changes. Abnormality, however, requires special handling: for subtractive changes, high abnormality is beneficial (removing unusual events), whereas for additive changes the measure is inverted to reflect the \q{normality} of the desired state. This ensures that abnormality is consistently treated as a beneficial property to be maximized during minimal-change computation.

To aggregate these four criteria, often conflicting, scores and select the optimal candidate, we employ the TOPSIS (Technique for Order of Preference by Similarity to Ideal Solution). TOPSIS is a well-established MCDM method chosen for its robustness and widespread use~\cite{taherdoost2023mcdm, hwang1981methods, taherdoost2023mcdm}, and for consistency with the framework we adapt for \foil determination~\cite{herbold2024contrastive}. The candidate ranked highest by TOPSIS constitutes the final minimal change presented in the explanation.

\begin{table*}
\centering
\caption{Desirable properties of minimal changes in counterfactual explanations}
\label{tab:desirable_properties}
\begin{tabularx}{\textwidth}{p{1.5cm} p{15.7cm} }
\toprule
\textbf{Property} & \textbf{Description $|$  Sources} \\
\midrule
%%%%
Controllability: &  The user's ability to implement the suggested changes. The changes are classified as \textbf{actionable} (direct user control), \textbf{mutable but non-actionable} (indirectly controllable via rules), or \textbf{immutable} (no control). %Controllability is treated as a primary filter during candidate selection, not as a cost factor in our minimality computation. The framework prioritizes actionable changes and recursively analyzes the controllability of preconditions for any mutable changes. 
 $|$ \cite{byrne2019CFEinXAI, karimi2021recourse, guidotti2018blackbox, poyiadzi2020face} \\ \midrule

%%%%

Sparsity: & The number of individual changes required to achieve the \foil. As users prefer short and simple explanations, lower sparsity is considered better. Our framework uses this property in two ways: first, it is a primary cost factor to be minimized during the computation of the minimal change. Second, it serves as a hard constraint, automatically excluding any potential explanation that requires more than three changes. | \cite{mothilal2020MLClassifiers, verma2024counterfactual,dai2023effect}\\ \midrule
%%%%
Temporality: & People tend to undo more recent events rather than distant ones. The temporality of a change is defined by the time elapsed since the relevant state last occurred; the shorter the interval, the more favorable the change. If no timestamp exists, a maximum value is assigned to ensure exclusion. Thus, recent changes are prioritized in minimal-change computation, while older ones are penalized. | \cite{miller1990temporality}  \\ \midrule
%%%%
Proximity & It counts the number of resulting system changes if all candidate changes were applied. The score is determined by simulating the downstream effects of a change, including any new rules that would subsequently fire or be preempted. The calculation differs based on the explanation type: for additive changes, effects are simulated forward from the new state, while for subtractive changes, the state prior to the prevented event is considered. Lower proximity values indicate more desirable minimal changes.  | \cite{mothilal2020MLClassifiers} \\ \midrule
%%%%
Abnormality & Based on the cognitive principle that users focus on altering unusual or exceptional events. This property measures the abnormality of a state based on its historical frequency. |  \cite{kahnemantversky1982abnormality, byrne2019CFEinXAI}  \\ 
\bottomrule
\end{tabularx}
\end{table*}
\subsection{Generation of the Counterfactual Explanation}
The final step in our framework is to translate the optimal set of minimal changes, identified by the ranking process, into a human-readable explanation. This set contains specific instructions for additive changes (e.g., \q{device d should have had state s}) and general instructions for subtractive changes (e.g., \q{device d should not have had state s}). 
To present this information to the user, we employ a natural language template. The template is composed of four key elements: (1) the \DI, (2) its \EST (the \foil), (3) the required additive changes, and (4) the required subtractive changes. To emphasize that the explanation refers to minimal changes that should have occurred, the pattern uses counterfactual conditional tense. The resulting template is shown below:
\begin{tcolorbox}[colback=white, colframe=black!75, boxrule=0.5pt, arc=2pt, left=6pt, right=6pt, top=4pt, bottom=4pt]
\emph{The Device of Interest} would be in the \emph{Expected State} if 
\textit{additive changes} had happened and \textit{subtractive changes} had not happened.
\end{tcolorbox}

For example, suppose Alice enters the living room and notices that the lamp is turned on, which surprises her. She expected the lamp to remain off, as she intended to keep the room dark (\EST = ``lamp off''), but instead observes the fact (\CST) that the lamp is on. 

The system analyzes the situation and finds that two \emph{Active \DR}. First,  \texttt{\textbf{[DR-1, Priority 4]:} If it is after 5 p.m., turn on the lamp}. Second, \texttt{\textbf{[DR-2, Priority 2]:} \emph{If the sun has set, turn on the lamp}}. Note that, while the only \DR with higher priority initially turned on the lamp, both remain active at the moment of explanation. 

To achieve the \foil (\EST = ``lamp off''), the system finds two \AR rules. First, \texttt{\textbf{[AR-1, Priority 1]:} If it is sunny, turn off the lamp}. Second, \texttt{\textbf{[AR-2, priority 3]:} If the room is empty, turn off the lamp}.  
The framework computes the minimal changes required for any (theoretically) possible \foil achievement:  
\begin{itemize}
    \item To \emph{Inactivate} \textit{DR-2}, that has a higher priority, the condition $\{ \text{sun\_set = false} \}$ must hold. To \emph{Override} \textit{DR-1}, either \textit{AR-1} or \textit{AR-2} must fire. 
    \item To override \textit{DR-2} directly, \textit{AR-1} must be fired, which requires the $\{ \text{weather = sunny} \}$.  
    \item Finally, the system can \textit{Inactivate} both DR-1 and DR-2 by setting $\{ \text{time = before 5 p.m., sun\_up = true} \}$.  
\end{itemize}  

All of these candidates go through the minimality computation. Among all, \q{making the room empty} is deemed minimal since having a \q{sunny weather} or \q{set time before 5 p.m.} 
are \textit{immutable}. The framework selects the minimum change set and generates the following explanation:  \textit{\q{The lamp would have been off if the room had been empty.}}

\subsection{Implementation}\label{sec:imple}
\begin{figure}[ht]
  \centering 
  \includegraphics[width=\columnwidth]{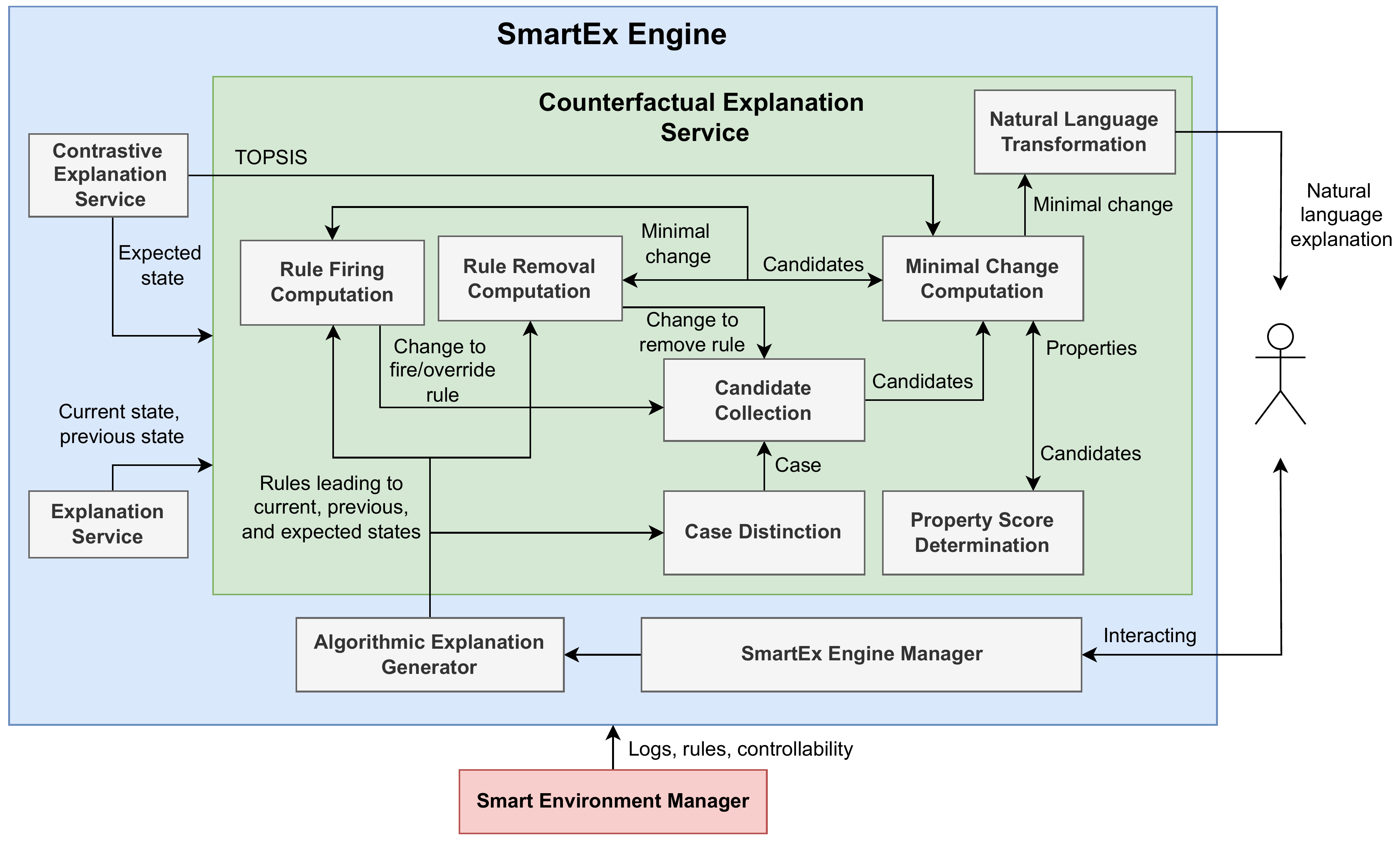}
  \caption{%
  	Reference architecture of \textit{SmartEx} and the \textit{Counterfactual Explanation Service}%
  }
  \label{fig: Reference Architecture}
\end{figure}

We implemented our framework as a plugin\footnote{\url{https://github.com/ExmartLab/SmartEx-Engine/tree/counterfactual}} for \textit{SmartEx} ~\cite{sadeghi2024smartex}, a RESTful Java service with MongoDB, integrated into smart environments via Home Assistant\footnote{\url{https://www.home-assistant.io/}}. \textit{SmartEx} provides causal~\cite{sadeghi2024smartex} and contrastive explanations ~\cite{herbold2024contrastive}, and our plugin extends it with counterfactual capabilities. Figure~\ref{fig: Reference Architecture} shows the references architecture of our framework.

When a counterfactual explanation is requested, \textit{SmartEx} identifies the \DI and fetches all relevant states (\CST, \PST) as well as all \emph{Active Rules} related to the \DI. It then interacts with the existing \textit{Contrastive Explanation} plugin for \foil and \EST determination.  

Given that, the \textit{Case Distinction Component} then identifies the explanation case (\exci--\exciii). The \textit{Candidate Collection Component} gathers all \DRs and \ARs, computes the best \foil achievement strategy, and creates combinations of how candidate rules can be \textit{Activated}, \textit{Inactivated}, or \textit{Overridden}. The \textit{Minimal Change Computation Component} removes duplicates and excludes all candidates that are not \textit{actionable} if fully \textit{actionable} ones exist. Finally, candidates are scored for sparsity, temporality, proximity, and abnormality. Using TOPSIS, the best candidate is selected and transformed into natural language. The generated explanation is finally delivered to the user through \textit{SmartEx}’s smartphone or web application.

\section{Evaluation}\label{sec:evaluation}
To assess how our generated counterfactual explanations are received in practice, we conducted a quantitative, human-centered evaluation. The study was designed to compare counterfactual explanations against traditional causal ones, guided by two research questions (RQ):
\begin{itemize}
    \item \textbf{{RQ1}:} Do users prefer counterfactual or causal explanations in smart environments?
    \item  \textbf{{RQ2}:} In which contexts do users prefer counterfactual or causal explanations in
   smart environments?
\end{itemize}

\subsection{Study Design}\label{sec:studyDesign}
We conducted a study using a within-subject experimental design with 17 participants, recruited via personal contacts. Each participant took part in a 15-minute, in-person interview session. After a brief welcome and introduction, participants were informed about data privacy and asked two preliminary questions to assess their general preferences: (1) short vs. detailed explanations, and (2) explanations that provide reasons vs. those that offer solutions. These baseline responses were later used to contextualize participants’ preferences (\hyperlink{rq2}{RQ2}).
\begin{table*}
    \centering
     %\footnotesize
    \caption{Explanations provided to participants}
    \label{tab: Provided Explanations}
    \begin{tabular}{@{}c p{2cm} p{14.5cm}@{}}
        \toprule
        \textbf{Scene} & \textbf{Exp. Type} & \textbf{Explanation} \\ 
        \midrule
        1 & Causal & The speaker is on because no meeting is going on in a meeting room, and the social room is not empty. \\
          & Counterfactual & The speaker would be off if there was a meeting going on in a meeting room. \\\midrule
        2 & Causal & The meeting room door is locked because it is before 8:30~a.m. \\
          & Counterfactual & The meeting room door would be open if it was not before 8:30~a.m. \\\midrule
        3 & Causal & The brightness is at 70\% because there is only a single person in the room. \\
          & Counterfactual & The brightness would be at 100\% if a device was connected to the beamer. \\\midrule
        4 & Causal & The speaker remains off because no rule was executed. \\
          & Counterfactual & The speaker would be on if there was no meeting going on. \\\midrule
        5 & Causal & The air conditioning is on because it is sunny and all windows are closed. \\
          & Counterfactual & The air conditioning would be off if the door was open longer than 10 minutes and not all windows were closed. \\\midrule
        6 & Causal & The blinds are rolled down halfway because the blind’s controller down button was pressed twice, and the plant lights are off. \\
          & Counterfactual & The blinds would be rolled down completely if the plant lights were not off. \\
        \bottomrule
    \end{tabular}
\end{table*}
Participants then experienced two narrative-driven scenarios comprising a total of six \textit{confusing scenes} caused by automation. Each scene was presented through a slide-based illustration of a 3D-style top-down room view, with clearly depicted devices (e.g., speakers, blinds, air conditioning), brief textual descriptions, and sound or visual cues (e.g., musical notes or brightness level changes) to support the user’s mental model and enhance immersion.

Before each scene, participants were reminded that all upcoming explanations were factually correct to avoid bias in evaluation. After encountering the confusing situation, they were asked whether they \textit{wanted} an explanation (choosing between \q{yes}, \q{I don't care}, or \q{no}). This question was included solely for analytical purposes but it did not affect the study procedure. Regardless of their answer, all participants were then provided with three paper snippets: a \textit{causal explanation}, a \textit{counterfactual explanation}, and a \textit{no explanation} option. They
were asked to rank these options based on their subjective
preference.

Scenes varied across several dimensions\footnote{Due to space limitations, we do not provide descriptions of the scenes here. Please refer to the extended version (\url{http://kups.ub.uni-koeln.de/id/eprint/78813}) for complete details.}: environmental setting (home vs. office), urgency (scenes with and without time pressure), and the degree to which \textit{causal and counterfactual explanations diverged}. In some cases, such as Scene 2, counterfactual explanations were concise and actionable, proposing specific changes to resolve the situation (Table~\ref{tab: Provided Explanations}). In contrast, in some scenes such as Scene 5, where no actionable change was possible (i.e., all relevant conditions were \textit{immutable}), the counterfactual and causal explanations were functionally equivalent, differing only in linguistic framing.

%\begin{quote}
%\textbf{Causal:} The brightness is at 70\% because there is only a single person in the room.\\
%\textbf{Counterfactual:} The brightness would be at 100\% if a device was connected to the beamer.
%\end{quote}

%\begin{quote}
%\textbf{Causal:} The meeting room door is locked because it is before 8:30 a.m.\\
%\textbf{Counterfactual:} The meeting room door would be open if it was not before 8:30 a.m.
%\end{quote}

After all six scenes were completed, participants were formally introduced to the distinction between \textbf{causal} and \textbf{counterfactual} explanations, and were provided with a reference list of all explanations encountered during the study. Finally, participants completed a 4-item Likert-scale questionnaire, rating each explanation type (causal and counterfactual) on (1) linguistic clarity and (2) content usefulness.

\subsection{Results and Discussion}
To investigate users’ preferences for different explanation types in smart environments, we analyzed their rankings and ratings across six interactive scenes.
\begin{figure}
    \centering
    \includegraphics[width=.47\textwidth]{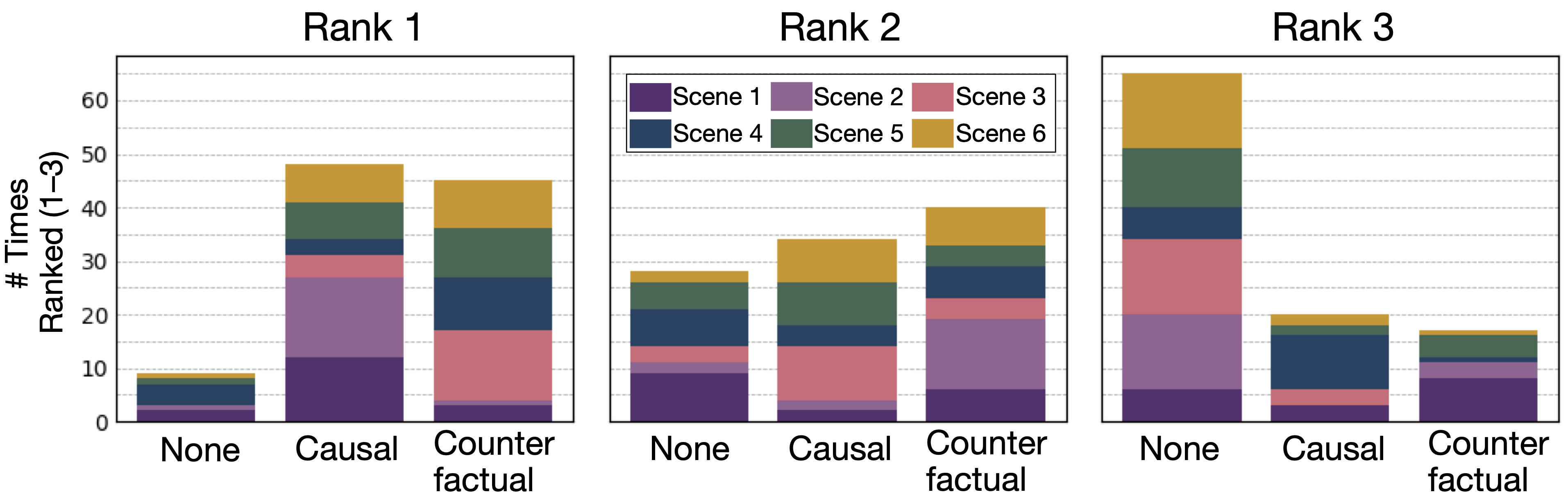}
    \caption{\small Distribution of explanation type rankings across six scenes and 17 participants. It shows how often each explanation type was ranked 1st, 2nd, or 3rd across all participant–scene combinations (total: 102 per rank).}
    \label{fig: combined rankings all}
\end{figure}
Regarding \textbf{RQ1}, we found no strong overall preference for either explanation type. As shown in Figure~\ref{fig: combined rankings all}, \textit{causal explanations were ranked first slightly more often} than counterfactual ones, but they were also ranked third more frequently, indicating more polarized opinions. Causal explanations were especially favored in the two initial scenarios, which involved time-sensitive or straightforward automation events. In contrast, counterfactual explanations were preferred in the remaining scenarios, particularly those without time pressure or where a suggested action could alter the system’s behavior. Across all scenes, the \q{no explanation} option was generally least preferred. This lack of a clear overall winner suggests that user preference is not absolute and is instead driven by specific contexts, which we explore in our analysis of RQ2.

To answer \textbf{RQ2}, we analyzed how preferences varied across different contexts:

\textbf{Explanation-Specific Factors:} The post-study questionnaire revealed that the evaluation criterion is a primary contextual factor. Participants expressed a strong preference for causal explanations \textit{linguistically} (Mean 4.24 vs. 2.94). However, when judging the \textit{content}, they rated counterfactual explanations slightly more favorably (Mean 3.65 vs. 3.53). This crucial distinction shows that while users find the language of causal explanations easier to parse, they often find the information within counterfactuals more valuable.

\begin{figure*}
    \centering

    \begin{subfigure}{0.24\textwidth}
        \centering
        \includegraphics[height=3.2cm]{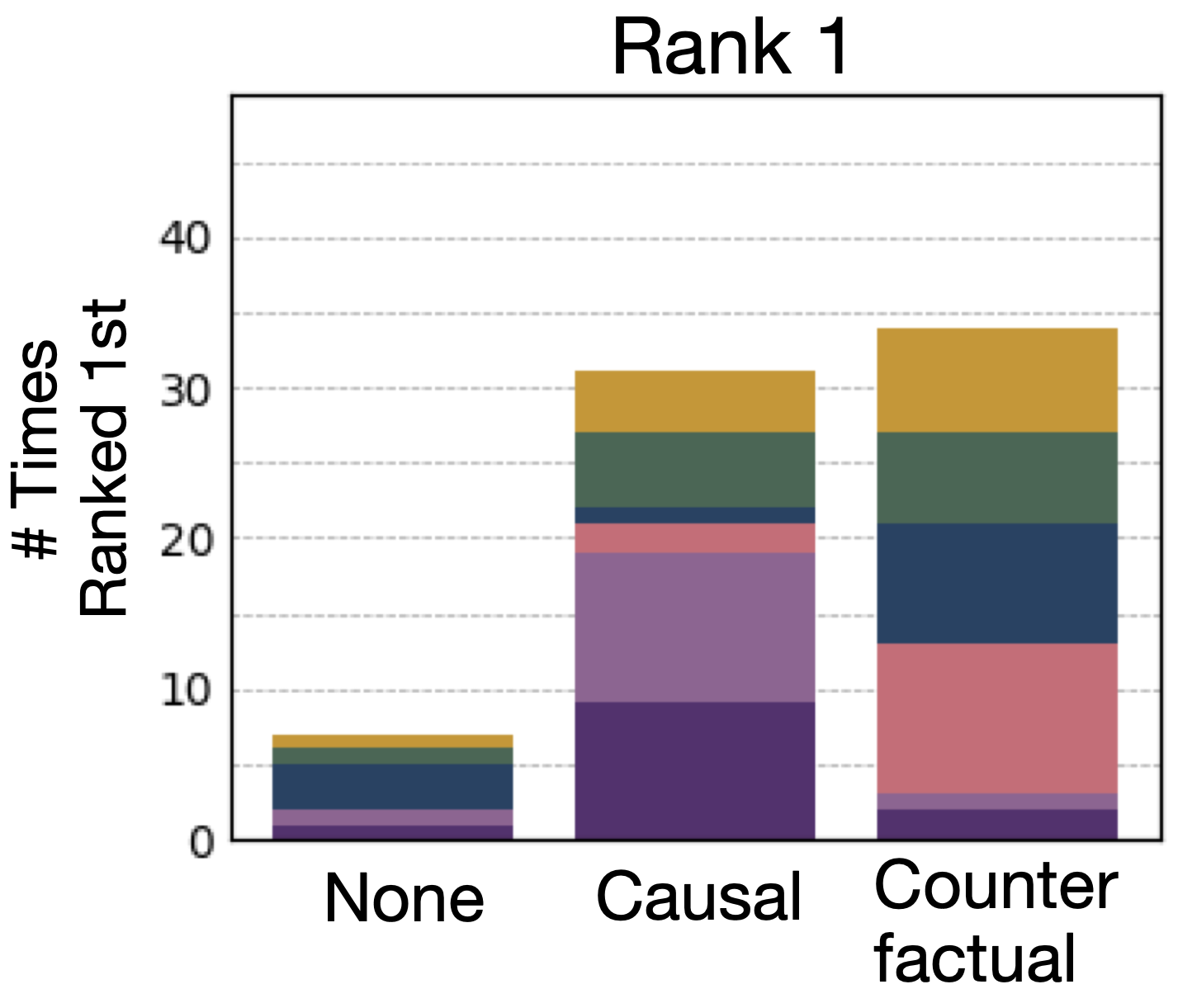}
        \caption{Shorter Explanation}
        \label{fig:preference_short}
    \end{subfigure}
    \hfill
    \begin{subfigure}{0.24\textwidth}
        \centering
        \includegraphics[height=3.2cm]{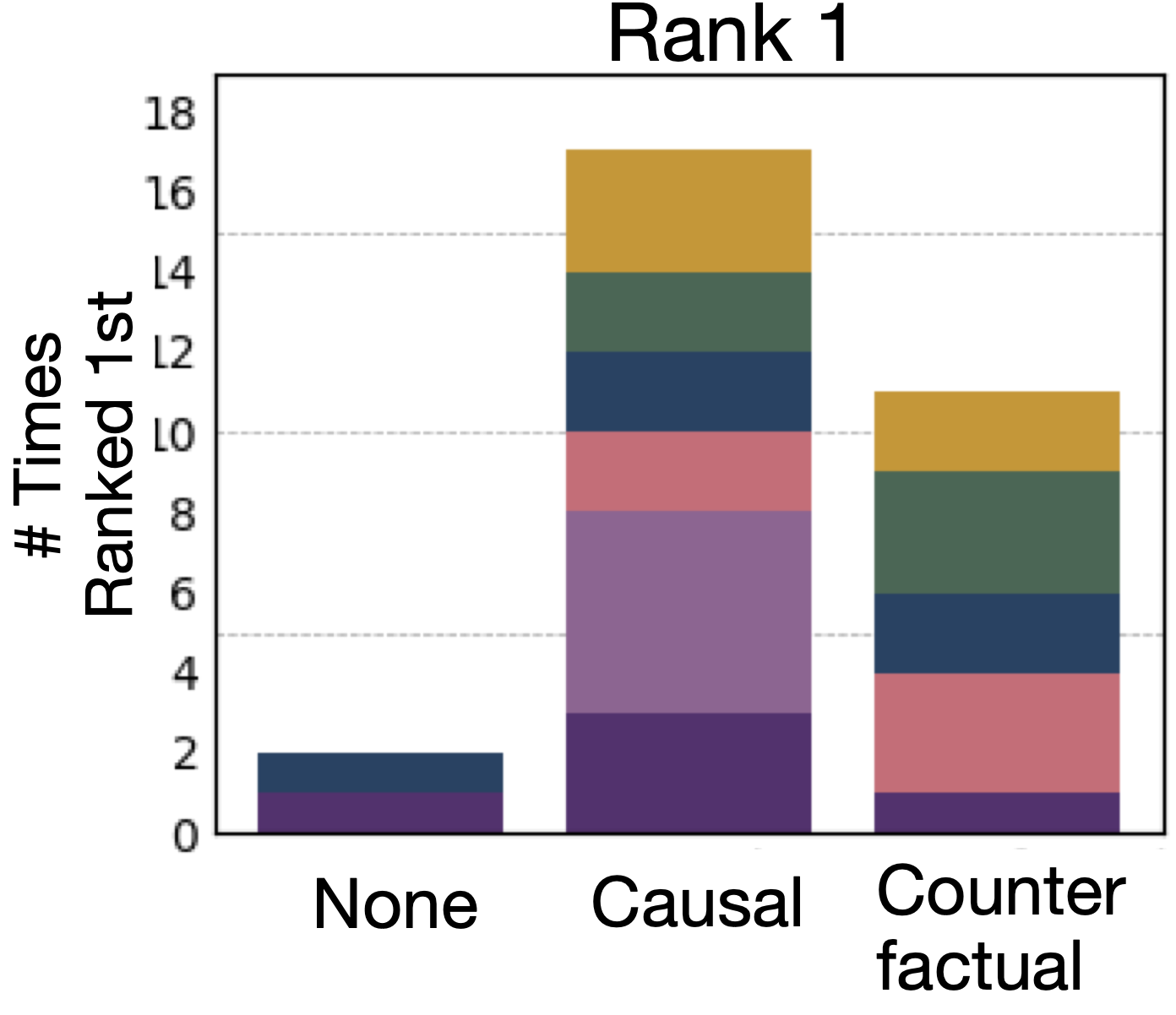}
        \caption{Longer Explanation}
        \label{fig:preference_long}
    \end{subfigure}
    \hfill
    \begin{subfigure}{0.24\textwidth}
        \centering
        \includegraphics[height=3.2cm]{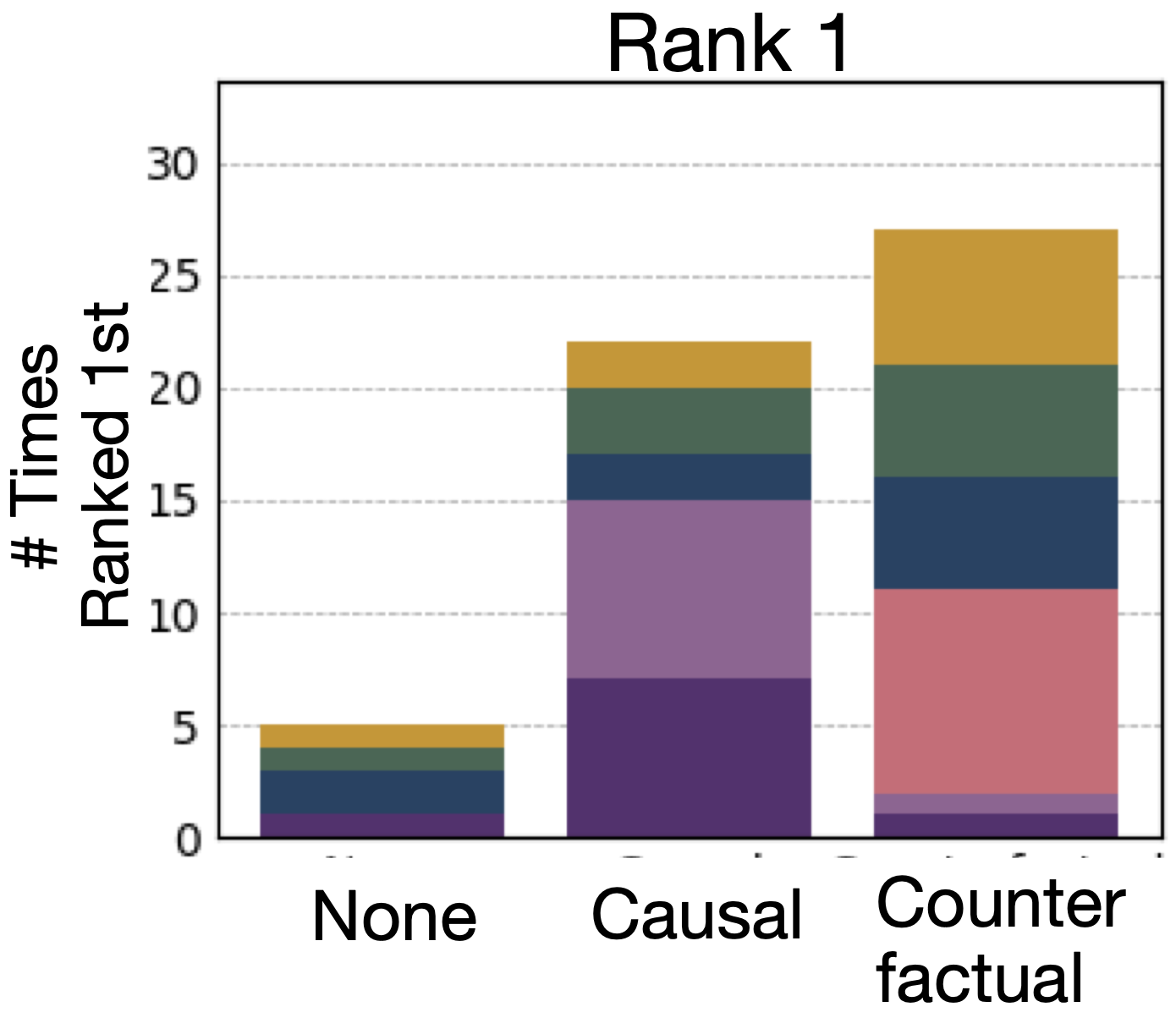}
        \caption{Solution Explanation}
        \label{fig:preference_solution}
    \end{subfigure}
    \hfill
    \begin{subfigure}{0.24\textwidth}
        \centering
        \includegraphics[height=3.2cm]{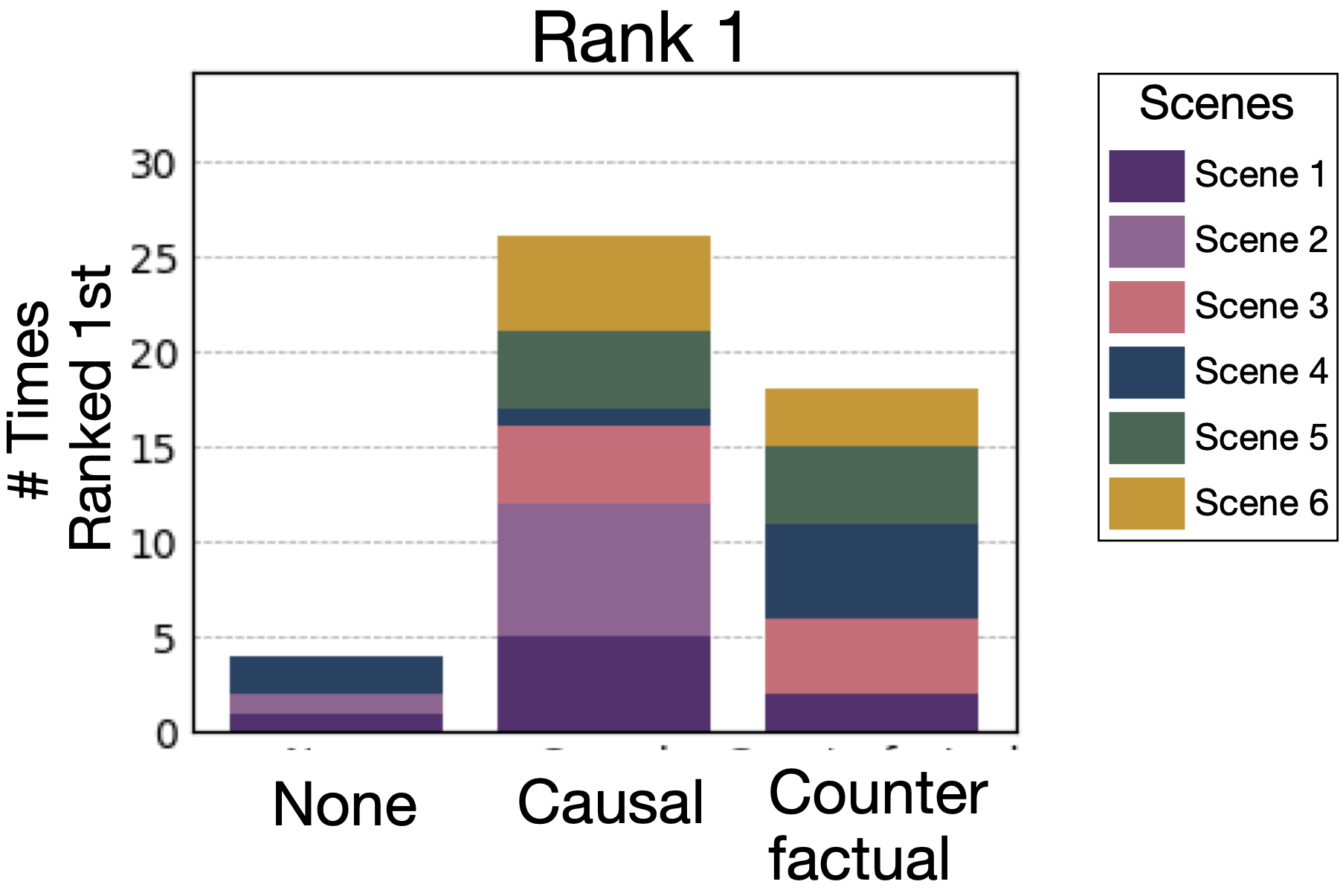}
        \caption{Reasoning Explanation}
        \label{fig:preference_reason}
    \end{subfigure}

    \caption{\small Number of times each explanation type was ranked 1st across all scenes, grouped by participants’ stated explanation preference.}
    \label{fig:combined_rankings_preferences}
\end{figure*}

\textbf{User's Goal and Style:} The user's preexisting preferences strongly correlated with their choices (see Figure~\ref{fig:combined_rankings_preferences}). Participants who prefer \textit{shorter, solution-oriented explanations} showed a clear preference for \textit{counterfactuals}. Conversely, those who prefer \textit{longer, reason-oriented explanations} strongly favored \textit{causal} ones.

\textbf{Situational Context:} The situation in each scene also had a significant impact. We found that \textit{causal} explanations were more preferred when users were under \textit{time pressure}, suggesting they require less cognitive effort to comprehend. In contrast, \textit{counterfactual} explanations were strongly preferred when users expressed a high \textit{need for an explanation} and a \textit{desire to change} the situation.

\textbf{Explanation-Specific Properties:} The actionability of the explanation was highly influential. The preference for counterfactual explanations was significantly stronger when they were \textit{actionable}, validating our framework's focus on this property. Furthermore, among the counterfactual explanations, those with \textit{subtractive structure}, emphasizing the inactivation of a triggering cause, were slightly more preferred than additive ones that introduced hypothetical alternatives. This may be because subtractive explanations align more closely with users’ intuitive causal reasoning and focus directly on what went wrong.

Overall, explanation preferences were context-dependent. Causal explanations were preferred in time-pressured or low-engagement contexts. Counterfactual explanations were valued when users desired control, actionable suggestions, or deeper understanding. These findings support the case for offering both types dynamically, tailored to situational and user-specific factors in smart environments.

\section{Conclusion}
In this paper, we addressed the lack of counterfactual explanations in rule-based smart environments by proposing a formal definition and a novel generation framework. Our approach operationalizes the principle of \q{minimal change} by scoring candidate explanations against five desirable properties (e.g., controllability, sparsity). We implemented and evaluated this framework in a human-centered study, which confirmed that user preference is highly contextual, revealing a trade-off between the linguistic simplicity of traditional causal explanations and the actionable, solution-oriented content of our counterfactuals. The key implication is that intelligent systems should adapt the explanation type to the user's context and goals rather than relying on a single method. While this study has limitations, our framework provides a foundation for future work, particularly in using large language models to improve the linguistic quality of counterfactuals and in conducting larger, in-situ validations.
%%REFERENCES:

\end{document}